\documentclass[letterpaper, 10 pt, conference]{ieeeconf}  

\usepackage{amsmath} 
\usepackage{amssymb}  
\usepackage{graphicx}
\usepackage{multirow}
\usepackage{makecell}
\usepackage{subfig}
\usepackage{balance}
\usepackage{color}
\usepackage{hyperref}

\usepackage{graphics} 
\usepackage{amsmath} 
\usepackage{amssymb}  
\usepackage[export]{adjustbox} 
\usepackage{hyperref}
\usepackage{cleveref}
\usepackage{mathtools}
\usepackage{bm}
\usepackage{tikz}
\usepackage{booktabs}
\usepackage{xcolor}
\usepackage{graphics}
\usepackage{multirow}
\usepackage{float}

\newcolumntype{L}[1]{>{\raggedright\let\newline\\\arraybackslash\hspace{0pt}}m{#1}}
\newcolumntype{C}[1]{>{\centering\let\newline\\\arraybackslash\hspace{0pt}}m{#1}}
\newcolumntype{R}[1]{>{\raggedleft\let\newline\\\arraybackslash\hspace{0pt}}m{#1}}

\newcommand{\norm}[1]{\left\lVert#1\right\rVert}
\newcommand{\R}{\mathbb{R}}

\IEEEoverridecommandlockouts                              

\title{\LARGE \bf
Learning Fine Pinch-Grasp Skills using Tactile Sensing from A Few Real-world Demonstrations
}

\author{Xiaofeng Mao, Yucheng Xu, Ruoshi Wen, Mohammadreza Kasaei, \\Wanming Yu, Efi Psomopoulou, Nathan F. Lepora, Zhibin Li
\thanks{Xiaofeng Mao, Yucheng Xu, Mohammadreza Kasaei and Wanming Yu are with the School of Informatics, University of Edinburgh, UK.
Ruoshi Wen is with the Touchlab Limited, Edinburgh, U.K.
Efi Psomopoulou and Nathan F. Lepora are with the Department of Engineering Mathematics, University of Bristol, UK. Zhibin Li is with the Department of Computer Science, University College London, UK. Corresponding author's email: xiaofeng.mao@ed.ac.uk}
}

\begin{document}

\maketitle
\thispagestyle{empty}
\pagestyle{empty}

\begin{abstract}
Imitation learning for robot dexterous manipulation, especially with a real robot setup, typically requires a large number of demonstrations.
In this paper, we present a data-efficient learning from demonstration framework which exploits the use of rich tactile sensing data and achieves fine bimanual pinch grasping. Specifically, we employ a convolutional autoencoder network that can effectively extract and encode high-dimensional tactile information. Further, We develop a framework that achieves efficient multi-sensor fusion for imitation learning, allowing the robot to learn contact-aware sensorimotor skills from demonstrations. Our comparision study against the framework without using encoded tactile features highlighted the effectiveness of incorporating rich contact information, which enabled dexterous bimanual grasping with active contact searching. Extensive experiments demonstrated the robustness of the fine pinch grasp policy directly learned from few-shot demonstration, including grasping of the same object with different initial poses, generalizing to ten unseen new objects, robust and firm grasping against external pushes, as well as contact-aware and reactive re-grasping in case of dropping objects under very large perturbations. 
Furthermore, the saliency map analysis method is used to describe weight distribution across various modalities during pinch grasping, confirming the effectiveness of our framework at leveraging multimodal information. 
The video is available online at: \href{https://youtu.be/BlzxGgiKfck}{https://youtu.be/BlzxGgiKfck}.

\end{abstract}


\section{INTRODUCTION}


Dexterous robot manipulation has the capability to work across a range of tasks and environments. However, enabling dexterous manipulation in robots, particularly in a manner that is comparable to human capabilities, remains an unsolved challenge. Currently, numerous studies utilize visual feedback to enable robots to perform dexterous manipulation tasks such as box flipping~\cite{zhu2019dexterous}, object rotating~\cite{arunachalam2022dexterous}, and door opening~\cite{qin2023dexpoint}. However, these visual-based methods have limitations, as the visual data could be influenced by occlusion and lighting variations. Consequently, it is very important to investigate how to incorporate tactile information for the enhancement of dexterous manipulation in robotic systems.

Tactile sensing plays a vital role in capturing detailed information about contact surfaces, including the distribution of contact forces and their variations during force-sensitive tasks -- which is indispensable for achieving dexterous handling of lightweight objects with irregular surfaces, shapes, and deformable properties. Especially during close-range interaction between hands and objects, visual occlusion restricts the ability to perceive detailed information of the contact surfaces, during which tactile sensors become valuable for providing essential information of these unseeable surfaces. Integrating tactile sensing into motor learning of dexterous grasping can enhance the rich and precise sensing of surface contacts and interaction dynamics, provide irreplaceable and direct feedback when manipulating objects, and enable more robust and precise manipulation tasks~\cite{li2020review, lepora2021soft}. It is crucial to explore how robots can leverage this information to achieve dexterous manipulation abilities. 

\begin{figure}[t]
    \centering
	\includegraphics[trim=0cm 0cm 0cm 0cm,clip,width=\linewidth]{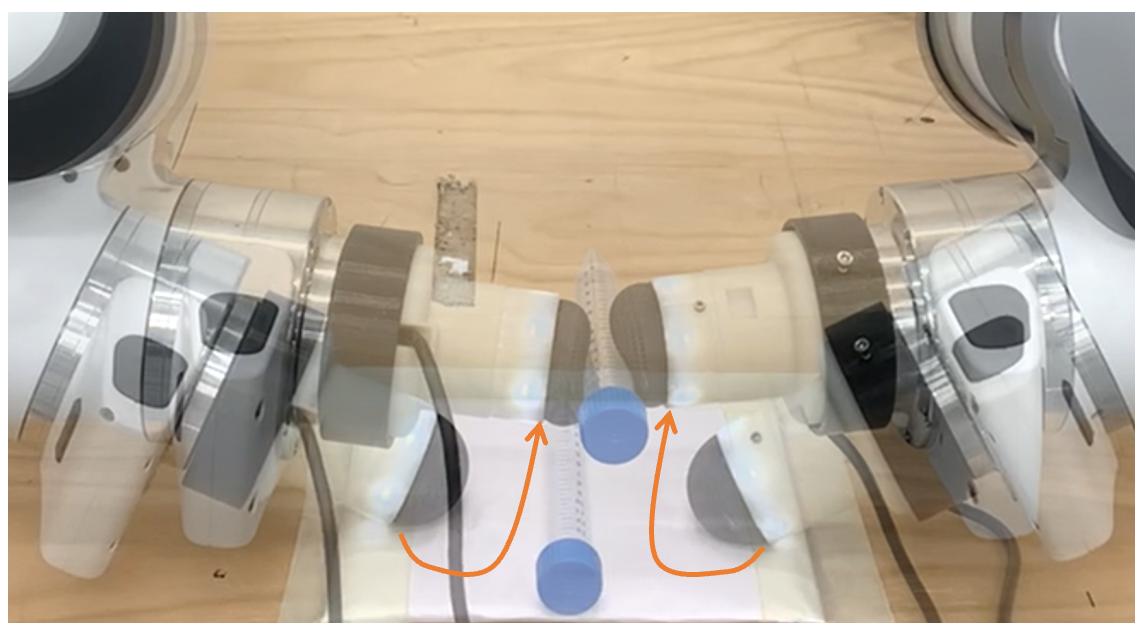}	
    \caption{Autonomous dexterous grasping with soft tactile sensors, including pre-grasp, press, roll-lift, and firm grasp.}
    \label{fig:cover}
    \vspace{-5mm}
\end{figure}

The canonical hardware for robot manipulation incorporates Force/Torque sensors that can only measure the 6-degree-of-freedom (DoF) wrench at each end-effector. Soft optical-based tactile sensors can provide abundant and discriminative contact information by quantifying the deformation of the soft materials using a camera system~\cite{lepora2021soft}. 
Currently, several soft tactile sensors have been developed, including TacTip~\cite{TacTipFamily2018}, DigiTac~\cite{lepora2022digitac}, Gelsight~\cite{GelSight2017}, and DIGIT~\cite{lambeta2020digit}. However, how to use high-dimensional data from tactile sensors for robot contact-rich tasks remains open research. 




The complex and non-trivial deformation of soft tactile sensors during dexterous grasping tasks presents a considerable challenge. Humans can deal with soft contacts, quickly adapt to new tasks, and produce skills of dual-arm coordination for manipulating objects. Learning from Demonstration (LfD) offers an intuitive, efficient method for acquiring human skills through synchronized tactile information, encoding rich state-action mapping and enabling robots to learn human sensorimotor skills while responding to tactile and proprioceptive feedback. Additionally, the issue of errors accumulating in contact-rich tasks during Learning from Demonstration (LfD), due to the lack of direct feedback, can be addressed by incorporating rich tactile feedback in real-time. The challenge involves effectively fusing high-dimensional data with robot proprioceptive states for sample-efficient human dexterous manipulation behavior learning. 

\subsection{Contribution}
In this work, we present a framework to teach bimanual, dexterous sensorimotor skills. To handle the complex dynamics involved with deformable tactile sensors, we employ behavior cloning (BC) to learn from human teleoperated demonstrations. A convolutional autoencoder (CAE) is trained in a self-supervised manner to extract essential features from the rich tactile data, which are then integrated with the robot's proprioceptive state. This multimodal fusion enables the robot to efficiently acquire feedback-driven dexterous grasping skills through a few human demonstrations. 
The proposed framework is validated by pinch grasp tasks on a dual-arm setup equipped with TacTips sensors~\cite{TacTipFamily2018} and has achieved the successful retrieval of a small, cylindrical object on a table using few-shot demonstrations. Our experimental results show that the policy, learned from few-shot human demonstration data, can achieve stable grasping of unseen objects with different diameters, masses, and materials. 

Furthermore, the robustness of the framework against external disturbances has been validated, with the learned policy demonstrating stable grasping under external disturbance, as well as the capacity to autonomously execute successful re-grasping in case of a large external force that pushes off the object. 
We applied saliency map analysis~\cite{simonyan2013deep} and revealed how the learned policy uses different sensory modalities in a variable way throughout the dexterous pinch grasp process. This analysis demonstrates the capability and effectiveness of our proposed network to efficiently use high-dimensional data and autonomously segment the long-horizon data into several distinct fine-skills for execution according to different contact situations.

\section{Related Works}
\label{sec:relatedwork}



During robotic dexterous manipulation, tactile sensors can provide rich contact information which is not easily accessed via visual information, thereby playing a crucial role in enhancing the dexterous grasping capabilities~\cite{jiang2022robotic}.
Soft deformable tactile sensors can perform contact-rich interactions with the environment and manipulate delicate objects safely~\cite{oller2023manipulation}. With optical-based tactile sensors, the orientation of the contact surface can be inferred from the tactile image, enabling stabilization of the pinch grasp by rolling the sensor on the contact surface and applying desired grasping forces~\cite{psomopoulou2021robust}. The study in~\cite{lin2023bi} explores utilizing tactile feedback to achieve bimanual tasks including bi-pushing, bi-reorienting, and bi-gathering, conducted within a simulated environment using reinforcement learning, and successfully achieves simulation-to-reality transfer. In contrast, our work focuses on using both deformable properties and rich surface contact information provided by the tactile sensor to achieve dexterous bimanual pinch grasping of small, fragile, and delicate objects.

One open question with high-dimensional tactile sensors is how to extract useful information from them. The works in~\cite{lepora2019pixels} estimate 6D contact wrenches from tactile images and the estimated wrenches that can be used as feedback to the grasping controllers within the classical control theory. Deep neural networks can also be used to process tactile images. The works in~\cite{lloyd2021goal} show that contact poses can also be detected from tactile images, which was then combined with goal-driven methods to achieve non-prehensile object stable pushing. The works in~\cite{TactileConvAE2019} introduce Autoencoder networks~\cite{rumelhart1985learning} to compress the high-dimensional tactile images into low-dimensional latent vectors which can be used for several down-stream tasks, such as object classification. In our work, we similarly employ an autoencoder to extract and encode tactile images. However, we extend its utility to achieve data-efficient learning from human demonstration by fusing the encoded tactile feature with robot proprioceptive information.



Moreover, although deformable tactile sensors facilitate area contact, potentially improving grasp stability and protecting delicate objects, the dynamics of the deformable sensor cannot be neglected. The work proposed in ~\cite{oller2023manipulation} combines 3D geometry of the tip of a deformable tactile sensor with robot proprioceptive action to learn the tactile sensor membrane dynamics and predict the deformation conditioned on robot action. Data-driven method can be used to learn the dynamics and combined with the Model Predictive Control (MPC) methods to achieve tactile servoing~\cite{tian2019manipulation}. Insights from human intrinsic understanding may prove valuable in leveraging deformable sensors to achieve dexterous dual-arm manipulation tasks. LfD is an intuitive and effective way to learn human skills from collected demonstrations, which is very helpful for tasks requiring high-level skills, such as intricate coordination between two arms. By segmenting the collected motion data, the work proposed in~\cite{caccavale2017imitation} generates a set of motion primitives to complete tasks. Additionally, human can use their senses to accomplish different tasks this can be used to investigate how the multi-sensory data can jointly together and help with the manipulation task~\cite{li2022see}.

\begin{figure*}[t]
    \centering
 	\includegraphics[trim=0cm 0cm 0cm 0cm,clip,width=\linewidth]{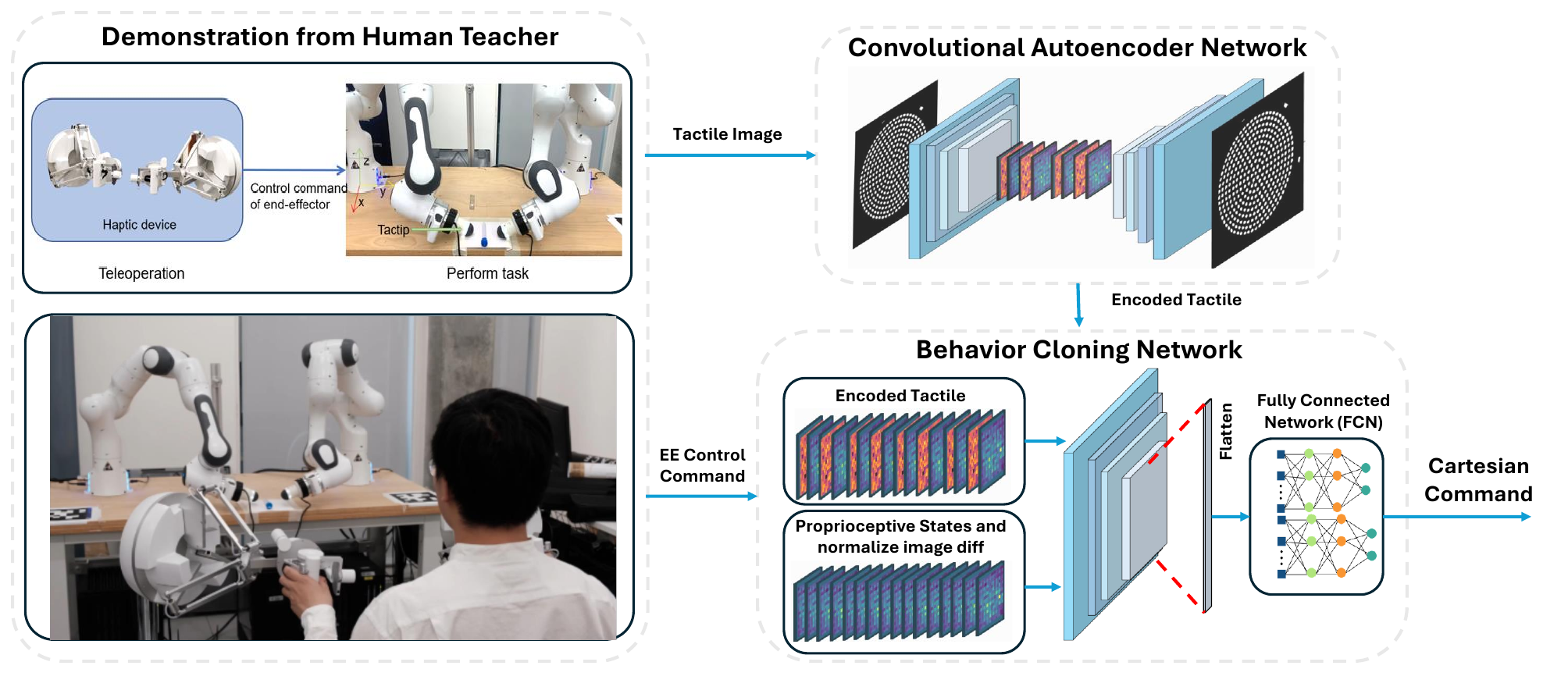}	 
    \caption{Architecture detailing the teleoperation system for demonstrations and the LfD framework.}
    \label{fig:overview}
    \vspace{-2mm}
\end{figure*}
\section{Methods}

    

\subsection{System Overview}
Teleoperation through a physical robot is a viable approach for generating real demonstration data that can be executed on a physical system, and it was shown to be effective in performing fine dexterous grasping \cite{wen2020force}. As shown in Fig.~\ref{fig:overview}, the overall architecture incorporates a teleoperation system for the collection of human demonstration data and a dual-arm setup for executing pinch grasp tasks. The teleoperation system consists of two haptic devices (Force Dimension Sigma 7) for human operators to control the dual-arm robot \cite{CollaboBimanual}. The dual-arm robot system includes two Franka Emika Panda arms each with a TacTip~\cite{TacTipFamily2018} installed on the end-effector of each arm. The Tactips capture contact information between end-effector and objects as 2D tactile images. Task-Space Sequential Equilibrium and Inverse Kinematics Optimization (SEIKO) runs in the backend to guarantee the guarantee of physical constraints and safety of the dual-arm robot \cite{rouxel2022multicontact}. 

The Learning from Demonstration (LfD) framework (see Fig.~\ref{fig:overview}) is composed of two distinct networks: 1)~a Convolutional AutoEncoder (CAE) network to extract the latent features from tactile images;~2) a Behavior Cloning~(BC) network to learn the policy of dexterous dual-arm grasping with tactile sensing from human demonstrations.

\subsection{Demonstration Dataset of Bimanual Manipulation} 
In our implementation, the haptic devices allow operators to adjust the 6D pose simultaneously, providing an intuitive way to demonstrate bimanual grasping skills on a dual-arm robot. During the demonstration, a human operator teleoperates the dual-arm robot to complete the grasping task by sending Cartesian commands to the two end-effectors via two haptic devices. The human demonstration data are recorded automatically during the entire grasping.

\subsection{Tactile Feature Extraction}
\label{sec:tactile}
The Tactip used in this work is an optical tactile sensor with a soft hemispherical tip, which was 3D-printed in one piece combining an elastic skin with 330 rigid white markers (pins)~\cite{TacTipFamily2018}. When the soft tip deforms during contact with objects, the white pins start to move away from their initial positions. The displacement of these pins reflects the complex deformation of the soft surface. An inner camera captures and projects the displacement to an array of white pins on a black background in the image plane. Raw tactile RGB images are firstly resized to 256$\times$256 pixels using linear interpolation and converted to grayscale images, which are then cropped using a circle mask and converted to binary images by thresholding. A median filter is applied to denoise the binary images.

We propose to use a self-supervised learning method -- convolutional autoencoder network to extract robust features that can represent the contact properties from the preprocessed tactile images. Eight convolutional layers are used in the CAE network to extract the spatial information represented by the displacement of the pins. The CAE network consists of an encoder and a decoder, formulated as follows: 

\begin{equation}
\label{eq:ConvEncoder}
\begin{array}{lc}
    g_\Theta(\cdot): \mathcal{X} \xrightarrow{} \mathcal{H}\\
    f_\Phi(\cdot): \mathcal{H} \xrightarrow{} \hat{\mathcal{X}}
    \end{array}.
\end{equation} 

The encoder~$g_\Theta(\cdot)$ projects each tactile image $\gamma_t$ in the high-dimensional input space~$\mathcal{X}$ (256$\times$256) to 16 feature maps $\gamma_l$ in the low-dimensional latent space~$\mathcal{H}$ (16$\times$16), then the decoder~$f_\Phi(\cdot)$ reconstructs that image from the same feature maps to the output space~$\hat{\mathcal{X}}$ (256$\times$256). The binary cross-entropy loss function is used as the reconstruction loss between the input images~$\mathcal{X}$ and the reconstructed images~$\hat{\mathcal{X}}$ to 
update the network parameters via back-propagation:
\begin{equation}
    \label{eq:CAE_loss}
    \begin{array}{cc}
      L_{CAE}(\gamma_t, \gamma_p) = -(\gamma_t\log\gamma_p) + (1-\gamma_t)\log(1-\gamma_p) \\
      \gamma_l = g_\Theta (\gamma_t), \gamma_p = f_\Phi(\gamma_l)
    \end{array},
\end{equation}
where $\gamma_p$ is the reconstructed image by the decoder network.

\subsection{Behavior Cloning Network}

We propose and design a BC network to learn the behaviors of coordinated manipulation skills of bimanual grasping from human demonstration data. Dexterous bimanual grasping skills can be considered into two categories: (1)~adaptive interaction with objects, and (2)~dual-arm motion coordination. To capture these skills, we have designed the input to our network to include encoded tactile feature maps, tactile image differences, and the robot's proprioceptive state.  The encoded feature maps and tactile image differences capture the human-object interaction skills. The robot's proprioceptive state, on the other hand, offers insights into the coordination of movements between both arms. These inputs collectively serve to reflect the complexity and adaptability of dexterous grasping skills.

Following this idea, we use the encoded tactile feature maps $l_t$, the proprioceptive state $\phi_t$, and the tactile image difference $e_t$ as input to the BC network to represent and learn fine human skills. The discrete-time state-action pair set $G = \left\{(s_0, a_0), (s_1,a_1), ..., (s_t,a_t),...\right\}$ is created to train the BC network, where $s_t = (l_t, \phi_t, e_t)$ denotes the robot state and $a_t$ denotes the Cartesian commands of the two arms at time $t$. 

Using such data of multiple modalities as input to train a network requires a well-crafted embedding structure~\cite{baltruvsaitis2018multimodal}. A common way of fusing a 2D feature map and a 1D feature vector is to flatten the 2D feature map into a 1D vector and concatenate the flattened vector and the 1D feature vector~\cite{akbulut2023bimanual}. However, we found that the flattening projection results in the \textit{loss of spatial correlation of tactile information}. For fine dexterous pinch grasping of small objects, the spatial information provided by tactile images is essential in understanding the contact situation and the object being contacted. To preserve the spatial information of the encoded tactile feature maps, we specifically tile the proprioceptive state of robots and the tactile image difference to \textit{match} the dimension of the tactile feature maps, so as to keep the spatial information of the encoded tactile feature maps.

\begin{figure*}[t]
    \centering
    \setlength{\belowcaptionskip}{0cm}
    \subfloat[Robustness of the learned control policy against external disturbances.]{\includegraphics[width=\textwidth]{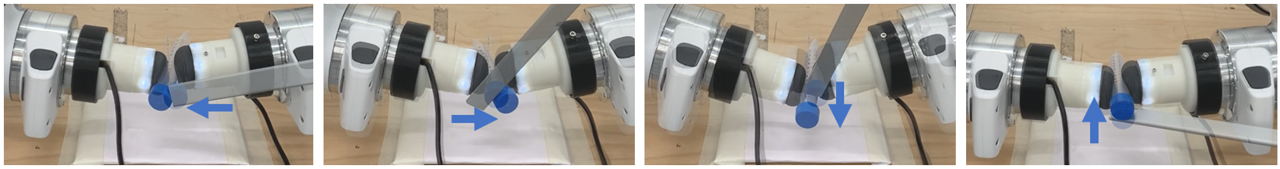}
    \label{fig:External_disturbance}} 
    \hfill
    \subfloat[Successful re-grasping using the learned policy.]{\includegraphics[width=\textwidth]{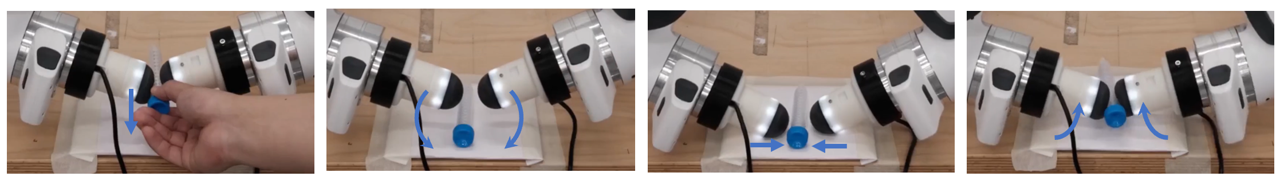}
    \label{fig:Regrasping}} 
    \hfill
    \subfloat[Successful grasping of unseen objects using the learned policy.]{\includegraphics[width=\textwidth]{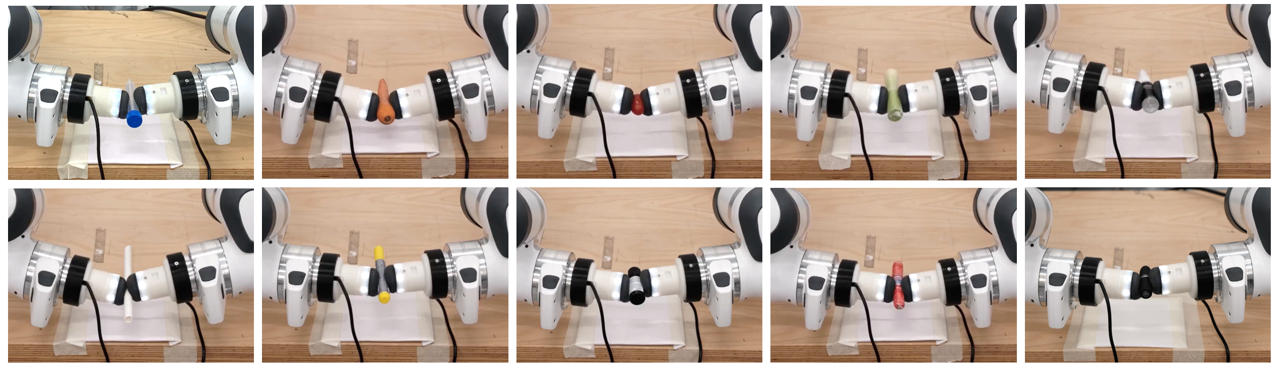}\label{fig:GENERALIZATION}} 
    \caption{Generalization of the learned policy and its robustness to external disturbance.}
    \label{fig:comparison}
\vspace{-5mm}
\end{figure*}

We then concatenate the tactile feature maps, the tiled proprioceptive state maps, and the tactile image difference on each feature channel. The convolutional layers in the BC network first filter the input feature maps (46$\times$16$\times$16) to a feature map (1$\times$8$\times$8), which is then flattened and fed into a fully connected network~(FCN). The FCN network outputs a vector $\hat{\bm{a}}\in \R^{12}$ as the predicted Cartesian pose commands of the two arms, including 3D position and 3D orientation for each arm.

The loss function used to train the BC network consists of two parts, which are formulated as: 

\begin{equation}
    \label{Loss Function}
    \begin{array}{cc}
         L_{BC}(\bm{a},\hat{\bm{a}}) = \norm{\bm{a} - \hat{\bm{a}}}^2 + \norm{\bm{d} - \hat{\bm{d}}}^2  \\
         
         \hat{\bm{a}} = \psi(l, \phi, e;\Phi_{conv}, \Phi_{fcn}) 
    \end{array}
\end{equation}
where $\bm{a}\in \R^{12}$ is the Cartesian pose commands of the two arms from the human demonstration dataset, and $\hat{\bm{a}}\in\R^{12}$ is the predicted Cartesian pose command by the BC network $\psi(\cdot;\Phi_{conv},\Phi_{fcn})$, parameterized by $\Phi_{conv}$ and $\Phi_{fcn}$; $l$, $\phi$ and $e$ denote the tactile feature maps, the proprioceptive state maps and the tactile image difference, respectively. The second term $\norm{\bm{d} - \hat{\bm{d}}}^2$ is added to learn the dual-arm coordination skills from human demonstrations, where $\bm{d}\in \R ^3$ is the relative position between the two end-effectors, and $\hat{\bm{d}}\in \R ^3$ is the predicted relative position between the two end-effectors by the BC network. 


\section{Experiments and Results}
\label{sec:experiments}
\subsection{Experimental Setup and Data Collection}

We validate the performance of LfD with tactile sensing for robot dexterous manipulation on the challenging task: the retrieval of small and fragile objects from the desk using dual-arm pinch grasp and and ensuring a stable grasp throughout the process.  During dexterous grasping, external vision can be easily occluded by the end-effector, potentially leading to inaccurate object estimation. Therefore, our experiments operate without using external visual sensors. By default, the starting position of the object lies between two robot hands, and the whole demonstration is operated in the task space. 


 We collected 5 demonstrations for this task. The human demonstration dataset collected in the grasping task includes three main components: the Cartesian commands, the proprioceptive states, and the tactile feedback (i.e., tactile images provided by the TacTip sensors). The Cartesian commands and the proprioceptive states of the two arms are collected at a frequency of 1000~Hz. Two Tactips record the tactile image pairs at a frequency of 60~Hz. For each demonstration, about 1500 tactile images are recorded. Before using the collected dataset to train the networks, several pre-processing methods are used to process the raw data. The proprioceptive states of the two arms and the tactile images, collected at different sampling rates, are synchronized using a linear interpolation method to align their timestamps. A median filter is then applied to smooth the Cartesian commands $a_t$, i.e., the 6D poses of two end-effectors. For raw tactile images, the structural similarity index measure (SSIM)~\cite{wang2004image} is used to quantify the difference between the current frame and the original frame.

\subsection{Implementation detail}
Our proposed model is developed using PyTorch~\cite{paszke2017automatic}. For the training of the CAE, we utilized a dataset that comprised 20 trials, including demonstrations of random behavior unrelated to our experiment. This dataset was collected using our robot setup, with each trial yielding approximately 1500 images captured during the demonstration phases. The trained CAE exhibits a satisfactory reconstruction quality, with a Mean Squared Error (MSE) loss of 0.015 and a Structural Similarity Index Measure (SSIM) of 0.934. The model training process for CAE, which involved 100 iterations, was completed in approximately two hours using an NVIDIA 1080 GPU. In the case of the BC network, the model was trained using data from 5 collected demonstrations, with the training process involving over 1,000 iterations and taking approximately 5 minutes to complete.

\subsection{Design of Validation Tasks}

\subsubsection{Learning grasping vial}

The human demonstrator performs teleoperation of dual-arm robots to grasp a plastic vial (a test tube with $\Phi=15.65$mm) that is horizontally placed on the table. 
A Behaviour Cloning (BC) network is trained using the gathered demonstration data, and the trained policy is tested on dual-arm robots to validate its generalisation on unseen initial poses. During the evaluative phase, we positioned the test tube between the end-effectors to evaluate the performance of the learned policy given variations in the starting position, specifically alterations of up to $\pm20$ degrees and displacements of up to $\pm2$ centimetres in the objects' locations.


\subsubsection{Generalization to unseen objects}
To evaluate the generalisability of the trained policy to unseen objects with a variation of radius, weight, or even materials~(e.g., soft and fragile objects), a set of test experiments have been conducted using multiple objects of different radii ranging from $11.7$mm to $28.6$mm.

\subsubsection{Robustness against external disturbance}
We also validate the robustness of the trained policy against external disturbances. We applied random external pushes from the left, right, up, and down directions on the grasped object to test if the two arms can coordinate their end-effectors' poses to ensure the balance of the object.

\subsubsection{Re-grasping capability}
The re-grasping experiments are conducted to test if the trained controller is contact-awareness and can perceive the loss of contact with the object in order to make necessary adjustments according to the tactile feedback and react to grasping failures. After the successful normal grasping, we severely pushed the object away to break its static equilibrium, and the object dropped down between two end-effectors again.




\begin{figure*}[t]
    \centering
    \setlength{\belowcaptionskip}{0cm}
    \subfloat[Unsuccessful grasping of the baseline policy using exactly the same BC network trained with unchanged tactile feedback.]{\includegraphics[width=\textwidth]{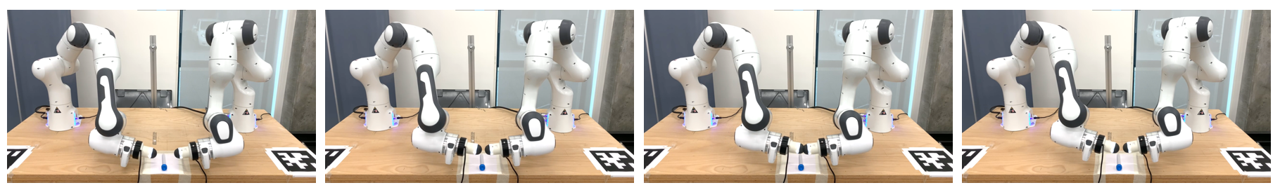}
    \label{fig:BC_freeze_image}} 
    \hfill
    \subfloat[Unsuccessful grasping of the baseline FCN policy trained with only the proprioperceptive information of end-effectors' pose and positions.]{\includegraphics[width=\textwidth]{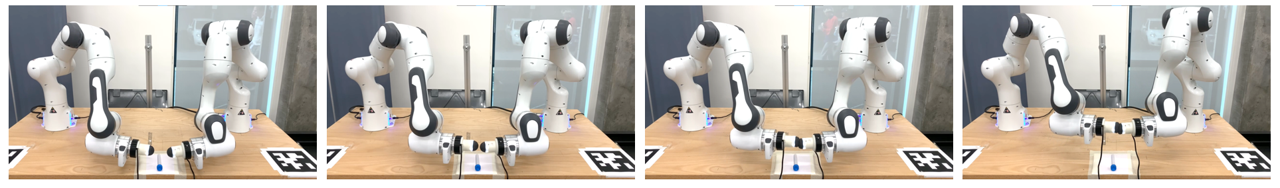}\label{fig:FCN}} 
    \caption{Results of the comparison study. The policy trained with both comparison frameworks bypass the object directly, manoeuvring the end-effectors directly to the desired end-poses without making any physical contact, grasping attempts, or interactions with the tube. }
    \label{fig:comparison}
    \vspace{-5mm}
\end{figure*}

\subsection{Results of Grasping Tasks}

The BC network trained on human demonstration data is deployed on a real dual-arm robot to verify its performance by the designed tasks. In both grasping tasks, the learned control policy achieved a $95\%$ success rate, even when the initial poses of the tube were different from their original pose in the demonstration. Occasionally, the policy might not succeed in grasping the object on the first attempt. However, it consistently adapts the behavior based on the contact situation with objects and will attempt to repeat the grasping action upon failure. The policy fails to grasp an object only when the object falls outside the workspace boundaries.

The dual-arm robot can make prompt adjustments and enable stable dexterous grasping by learning from only few demonstrations. In the process of lifting the object, the dual-arm robot achieves stable grasping by constantly twiddling the ``fingertips'' (tips of TacTip sensors) and adjusting the object to the central position. The process of retrieving an object from the table and adjusting its pose to maintain balance requires very fine movements and interactions supported by rich tactile information, where a 6-axis force/torque information is not sufficient to discern different contact situations in this scenario.


We evaluate the robustness of the learned policy against external disturbances. It can be seen from Fig.~\ref{fig:External_disturbance} that the dual-arm robot can make a proper adjustment to adapt to pushes. Although the pose of the two-arm robot in contact with the object was changed each time while being pushed, the dual-arm robot can always fine-adjust the object reactively to the center of the fingertips (Tactip sensors), roll and move the object to the desired position. Compared with the manually programmed behavior, this serves as a feedback policy that has been successfully acquired from human dexterity skills, which enables the dual-arm robot to autonomously adjust the posture and ensure a stable grasp quickly. It is noteworthy that such active rolling adjustment has not been specially demonstrated by any separate trials, but rather, this behavior was successfully captured by the rich tactile data during the demonstration of pick-lift grasping.

To examine the reaction in the presence of an unknown situation, i.e., grasping failures, the learned policy demonstrated contract-awareness of the falling object, i.e., loss of contact according to the tactile feedback, and thus controls the robot to restart the grasping process, which was not explicitly programmed or demonstrated by the prior LfD data. The result of the re-grasping experiments in Fig.~\ref{fig:Regrasping} shows that the tactile-based control learned from human demonstrations is very effective in performing robotic dexterous bimanual manipulation tasks autonomously and quickly without the need for explicit manual programming or complex planning.

The policy also achieves successful grasping of previously unseen objects, as shown in Fig~\ref{fig:GENERALIZATION}. Although the test objects have a variety of sizes and weights compared with the object used in the demonstration, the policy can still perform stable grasping. The experiment results show that the trained policy can generalize to unseen objects with similar cylindrical shapes but with different sizes and weights.

\begin{figure*}[t]
    \centering
    \setlength{\belowcaptionskip}{0cm}
    \subfloat[The output of the desired position and angles of the BC network for the dual-arm robot during dexterous grasping.]{\includegraphics[width=\textwidth]{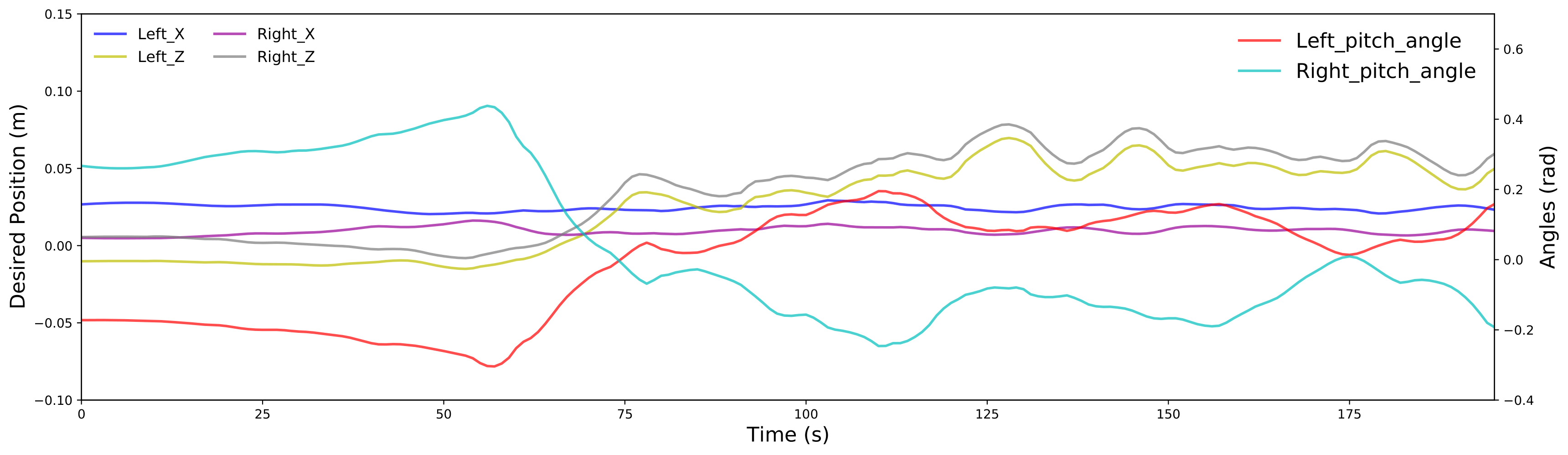}
    \label{fig:entire_grasping}} 

    \hfill
    \subfloat[Relative weight changes of each modality and its corresponding tactile information.]{\includegraphics[width=\textwidth]{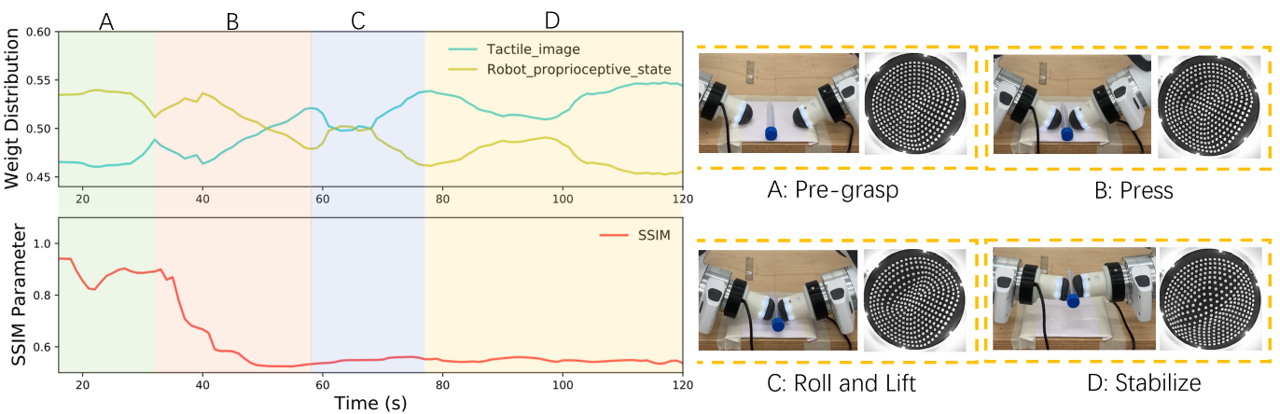}
    \label{fig:saliency}}
    \caption{The output of the learned policy and the weight changes during the grasping.}
    \label{fig:saliency_and_grasping}
\vspace{-5mm}
\end{figure*}


\subsection{Comparison Study}\label{sec:Comparison Study}

We conducted a comparison study to validate that successful grasping is achieved by the active use of tactile sensing. Besides training a BC network using the structure shown in Fig.~\ref{fig:overview}, we also train two different BC networks for comparison. The first one has exactly the same BC network structure but with frozen tactile images as input to CAE, meaning that the encoded image feature input stays unchanged during both the training and testing. The second one has an FCN structure and uses the poses of two end-effectors (both positions and orientation) as the input to train the network. 
The proposed BC network demonstrates convergence to a loss of 0.04 on the testing set. In contrast, the network employing frozen tactile information achieves convergence with a loss of 0.5, while the FCN converges to a loss value of 1.
These results prove that the effective integration of tactile information significantly enhances the convergence rate and leads to a reduced loss value in the final model.

We also compared all the grasping performances of the real dual-arm robot. As shown in Fig.~\ref{fig:comparison}, both BC network structures without using the tactile information failed in grasping the tube: robot arms failed to approach the object from its initial pose, and instead, they bypassed the object and moved towards the desired end-poses, showing no contact-awareness. The experimental results indicate that tactile feedback plays an essential role in providing contact information for initiating contacts, generating appropriate adjustments, lifting and retrieving to the desired target locations, enabling the dual-arm robot to perform very fine and dexterous contact-rich skills.

\subsection{Interpretability}
To explicitly show how much different modalities influence the entire operation, we use the saliency map method for calculating the weight distribution. The procedure for calculating this distribution is formulated as follows:

\begin{equation}
    \label{Saliency_equation}
    \begin{array}{cc}
     W_{i} = \frac{N(I)}{{N(I) + N(J)}}, W_{j} = \frac{N(J)}{{N(I) + N(J)}}
    \end{array},
\end{equation}
where $W_i$ and $W_j$ are the weight distributions of each modality. $N(\cdot)$ represents the normalization process. $I$ is the importance of the tactile information that is calculated by adding all the absolute values of weight that the learned policy distributed to tactile features. $J$ is calculated in the same way by adding all the absolute values of weight that are distributed to robot proprioceptive state features.

The comprehensive process of dexterous pinch grasping can be subdivided into four primary stages: pre-grasp, pressing, rolling and lifting, and stabilization. Each of these stages utilizes tactile feedback in a distinct manner.
In Fig~\ref{fig:saliency}, the weight changes during the complete dexterous pinch grasping process are depicted. Initially, as the end-effector moves toward the objects without any contact deformation on the tactile sensor, the weight of the robot's proprioceptive state exceeds that of the tactile information. When the tactile sensor comes into contact with the desk and is prepared for a pre-grasp pose, the weight of the tactile information increases (stage A). As the end-effector move towards the object and initiates contact, the weight attributed to the tactile information increases, exceeding that of the proprioceptive state (stage B). During the roll and lift phase, the weight of the tactile information initially decreases, subsequently achieving equilibrium with the proprioceptive state (stage C). This indicates that during the lifting phase, the learned policy necessitates both tactile information for successful in-hand manipulation and proprioceptive information for effective dual-arm coordination. Finally, upon successfully lifting the tube, the weight reverts to the tactile information, facilitating the stabilization of the tube (stage D).

\section{Conclusion And Future Work}
\label{sec:conclusion}
In this work, we introduce a tactile-driven LfD framework that demonstrates promising results in bimanual pinch grasping with a limited number of real robot demonstrations. Our exploration into leveraging the latest compliant tactile sensors has led to the development of the presented encoding methods that can effectively extract and capture high-dimensional contact sensing from soft tactile sensors, together with the fusion with proprioceptive feedback. The interesting outcome is to confirm the possibility of learning from real robot data directly, eliminating the necessity for large datasets and extensive training time, if the right data is effectively used. 

Our comparison studies showed that without the use of tactile sensing, dexterous motor skills cannot be learned by few-shot demonstrations with traditional robot sensing which is rather limited. In contrast, our approach demonstrates remarkable robustness in the presence of external pushes and is able to perform re-grasp the object if it drops. This ability was not explicitly illustrated in the initial demonstrations, emerging instead as a natural consequence of contact-aware sensorimotor skills through state-action mapping. 


Meanwhile, one apparent limitation is that the skill needs to be trained on a specific task, and it can be generalised and robust only around neighbourhood situations within a category of similar tasks: generalization applies to new/unseen objects that are similar to the demonstrated object of certain variations. Another limitation is that the robot's performance is based on blind grasping and re-grasping, and has not yet utilized external visual perception. In the future, integration of the current framework with stereo vision could extend the versatility and dexterity of object manipulation. Overall, our proposed LfD framework provides an attractive solution for learning from a few demonstrations with tactile sensing and supports broad real-world applications in contact-rich manipulation tasks.

\bibliographystyle{IEEEtran}
\bibliography{reference_mao}

\end{document}